\documentclass{article}
\usepackage{ijcai20}
\usepackage{times}
\usepackage{url}
\usepackage[small]{caption}
\usepackage{graphicx}
\usepackage{float}
\usepackage{tikz}
\usetikzlibrary{calc}
\usetikzlibrary{backgrounds,positioning}
\usetikzlibrary{decorations.pathreplacing}
\usepackage{adjustbox}
\usepackage{subcaption}
\usepackage{xcolor} 
\usepackage[linesnumbered,ruled,vlined,procnumbered,lined,boxed]{algorithm2e}
\usepackage{etoolbox}

\makeatletter
\AtBeginEnvironment{procedure}{\let\c@algocf\c@procedure}
\makeatother
\usepackage{multirow}
\usepackage{amsmath}

\DeclareMathOperator*{\argmin}{argmin}

\SetCommentSty{mycommfont}

\title{C-CoCoA: A Continuous Cooperative Constraint Approximation Algorithm to Solve Functional DCOPs}

\author{
Amit Sarker$^1$
\and
Abdullahil Baki Arif$^2$\and
Moumita Choudhury$^3$\And
Md. Mosaddek Khan$^4$
\affiliations
Department of Computer Science and Engineering, University of Dhaka
\emails
\{amitcsedu99$^1$, ahb.arif$^2$, moumitach22$^3$\}@gmail.com,
mosaddek@du.ac.bd$^4$
}

\begin{document}
\maketitle
\begin{abstract}
  Distributed Constraint Optimization Problems (DCOPs) have been widely used to coordinate interactions (i.e.\:constraints) in cooperative multi-agent systems.\:The traditional DCOP model assumes that variables owned by the agents can take only discrete values and constraints' cost functions are defined for every possible value assignment of a set of variables.\:While this formulation is often reasonable,\:there are many applications where the variables are continuous decision variables and constraints are in functional form.\:To overcome this limitation,\:Functional DCOP (F-DCOP) model is proposed that is able to model problems with continuous variables.\:The existing F-DCOPs algorithms experience huge computation and communication overhead.\:This paper applies continuous non-linear optimization methods on Cooperative Constraint Approximation (CoCoA) algorithm. We empirically show that our algorithm is able to provide high-quality solutions at the expense of smaller communication cost and execution time compared to the existing\:F-DCOP\:algorithms. 
\end{abstract}
\section{Introduction}
Distributed Constraint Optimization Problems (DCOPs) are a powerful framework to model cooperative multi-agent systems wherein multiple agents communicate directly or indirectly with each other. The agents act autonomously in a common environment in order to optimize a global objective which is an aggregation of their corresponding constraint cost functions. Each of the functions is associated with a set of variables controlled by the corresponding agents. In DCOPs, agents need to coordinate value assignments to their variables in such a way that maximize their aggregated utility or minimize the overall cost ~\cite{modi:adopt,petcu:scalable}. A number of multi-agent coordination problems, such as meeting scheduling \cite{maheswaran2004taking}, multi-robot coordination \cite{yedidsion2016applying} and smart homes \cite{fioretto2017multiagent,rust2016using}, have been dealt with this model.

The DCOP model is based on an assumption; that is, each of the variables that are involved in the constraints can take values from discrete domain(s) and a constraint is typically represented in a cost (i.e.\:utility) table. Nevertheless, a number of applications, such as target tracking sensor orientation \cite{fitzpatrick2003distributed}, cooperative air and ground surveillance \cite{grocholsky2006cooperative}, Network coverage using low duty-cycled sensors \cite{hsin2004network} and many others besides, can be best modeled with continuous-valued variables. Therefore, the traditional DCOP setting is not well-suited to such algorithms. To address this, the regular DCOP model is extended for continuous-valued variables \cite{stranders2009decentralised}. Later, \cite{hoang2019new} refer this continuous version of DCOP as Functional DCOPs (F-DCOPs). 

In more detail, \cite{stranders2009decentralised} propose a new version of the Max-Sum algorithm (i.e. Continuous Max-Sum\:-\:CMS) in order to solve continuous-valued DCOPs.\:CMS approximates constraint utilities as piece-wise linear functions. However, this approximation has not been widely recognised due to the unavailability of real-world applications having piece-wise linear functions.\:Then, Hybrid CMS (HCMS) uses discrete Max-Sum as the underlying framework with the addition of a continuous non-linear optimization method \cite{voice2010hybrid}. Notably, none of CMS and HCMS provides quality guarantees on the solutions as both of them are based on discrete Max-Sum which does not provide any quality guarantees when applied to general graphs \cite{hoang2019new}.\:To address this, three extensions of the Distributed Pseudo-tree Optimization Procedure (DPOP) \cite{petcu:scalable} algorithm has been proposed. The first one is an exact algorithm-Exact Functional DPOP (EF-DPOP) and the remaining two are non-exact methods $-$ Approximate Functional DPOP (AF-DPOP) and Clustered AF-DPOP (CAF-DPOP) \cite{hoang2019new}. EF-DPOP can solve F-DCOPs with tree-structured graphs and with linear or quadratic utility functions. AF-DPOP and CAF-DPOP can solve F-DCOPs without imposing restriction on the graph structure. However, as they are based on DPOP, a key limitation of these approximate algorithms is that they require exponential memory.

Against this background, we extend the Cooperative Constraint Approximation (CoCoA) \cite{van2017cocoa} algorithm so that it can solve functional DOCPs. We choose CoCoA as it is a non-iterative, semi-greedy approach that is able to find high-quality solutions with a smaller communication overhead than the state-of-the-art DCOP solvers. Our continuous version of CoCoA, that we call C-CoCoA, is an approximate local search algorithm that can solve F-DCOPs without any restriction on the graph structure and with a very lower communication cost. In C-CoCoA, we combine the discrete CoCoA algorithm with continuous non-linear optimization methods. Our target is to improve on continuous optimization by using the CoCoA algorithm to make the initial choice less critical. We empirically show that C-CoCoA outperforms HCMS and AF-DPOP in terms of solution quality, number of messages and time.

\section{Background}
\label{sec:2}
In this section, we discuss the background which is necessary to completely understand our proposed algorithm. We first describe the traditional DCOP model and then F-DCOP model. We then discuss the CoCoA algorithm and the challenges we face to incorporate CoCoA with the F-DCOP model.

\subsection{Distributed Constraint Optimization Problems}
A DCOP is defined as a tuple $\left\langle A, X, D, F, \alpha\right\rangle$, where,
\begin{itemize}
  \item A = $ \left\{ a_1, a_2,. .., a_n \right\} $ is a finite set of agents.
  \item X = $ \left\{ x_1, x_2,..., x_m \right\} $ is a finite set of discrete decision variables where each variable $x_i$ is controlled by one of the agents $a_i$ $\in$ $A$.
  \item D = $ \left\{ D_1, D_2,..., D_m \right\} $ is a set of finite discrete domains where each $D_i$ corresponds to the domain of variable $x_i$.
  \item F = $\{f_1, f_2,..., f_k\}$ is a finite set of cost functions, with each $f_i : \prod_{x_j \in x^i} D_j \rightarrow \!R$ defined over a set of variables $x^i \subseteq X$ and the cost $C$ for the function $f_i$ is defined for every possible value assignment of  $x^i$, that is, $C : D_{i_{1}} \times D_{i_{2}} \times...\times D_{i_{k}} \rightarrow \!R$.
  \item $\alpha:X \rightarrow A $ is a mapping function, which associates each variable $x_i \in X$ to an agent $a_i \in A$. An agent can control multiple variables. However, for simplicity, we assume each agent controls only one variable.
\end{itemize}
  A value assignment is complete if every variable is assigned a value. The goal in a DCOP is to find a complete assignment that minimizes the cost of the global objective function: 
  \begin{equation}
      X^* = \argmin_{X} \sum_{i = 1}^{k}f_i(x^i) \label{eq:1} \\
  \end{equation}
  
\subsection{Functional Distributed Constraint Optimization Problems}
\begin{figure}

  \begin{tikzpicture}
        [
        roundnode/.style={circle, draw=black, fill=white!5, very thick, minimum size=7mm},
        ]
        \node[roundnode]    at(1,0)  (x0)                              {$x_0$};
        \node[roundnode]    at(0,0)  (x3)                              {$x_3$};
        \node[roundnode]    at(2,1)  (x1)                              {$x_1$};
        \node[roundnode]    at(2,-1)  (x2)                              {$x_2$};

        \draw (x3) -- (x0);
        \draw (x0) -- (x1);
        \draw (x0) -- (x2);
        \draw (x1) -- (x2);
        \node  at (1,-3)
        {
            (a) Constraint Graph
        };
        \node at (5,1)  {
           $f(x_0, x_1) = x_{0}^{2} - 2x_0x_1$ + $2x_{1}^{2}$
        };
        \node at (5,0)  {
             $f(x_0, x_2) = x_{0}x_2 + 3x_{2}^2$
        };
        \node at (5,-1)  {
           $f(x_0, x_3) = x_{0}x_3 + x_{3}^2$
        };
        \node at (5,-2)  {
            $f(x_1, x_2) = x_{1}^{2} - x_1x_2 + 2x_{2}^{2}$
        };
        \node at (5,2)  {
             $D_i = [-20, 20]$
        };
        \node  at (5.5,-3)
        {
            (b) Cost Functions 
        };
\end{tikzpicture}
\caption{Example of an F-DCOP}
\label{fdcopex}
\end{figure}
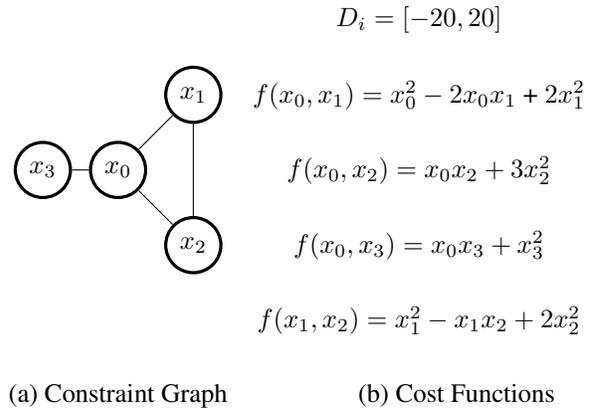

A Functional DCOP (F-DCOP) can be described by a tuple $\left\langle A, X, D, F, \alpha\right\rangle$, where A, F, and $\alpha$ are exactly the same as those in a DCOP. X and D are defined as follows:
\begin{itemize}
    \item X = $ \left\{ x_1, x_2,..., x_m \right\} $ is a finite set of continuous decision variables.
    \item D = $ \left\{ D_1, D_2,..., D_m \right\} $ is a set of continuous domains. Each variable $x_i$ can choose any value from a range, $D_i = [LB_i, UB_i]$.
\end{itemize}
As aforementioned in the previous section, the difference between F-DCOPs and DCOPs is found in the representation
of the cost function. In DCOPs, cost functions are represented in a tabular form. However, in F-DCOPs, we use a function to represent a constraint cost instead of the traditional tabular form. The goal of an F-DCOP is the same as a DCOP, which is finding a complete assignment that minimizes the cost of the global objective function. An example of an F-DCOP is presented in Figure~\ref{fdcopex} where Figure~\ref{fdcopex}(a) represents a constraint graph with four variables. Each variable $x_i$ is controlled by one of the agent $a_i$. The edges between the variables represent the cost functions that are defined in Figure~\ref{fdcopex}(b). The domain $D_i$ is defined as [-20, 20] in this example.

\subsection{Cooperative Constraint Approximation (CoCoA)}
\label{sec:cocoa}
The CoCoA algorithm starts with randomly activating an agent. Upon activation, the agent sends an inquiry message to its neighboring agents. We define the set of direct neighbors of the agent $a_i$ is $\mathcal{N}_i$. When an agent $a_i$ sends an inquiry message to the neighboring agents $a_j \in \mathcal{N}_i$, each $a_j$ calculates cost messages for every value in the domain of $a_i$ using Equation~\ref{eq:21}. Here, $\zeta_{j,k}$ is the cost for the $k^{th}$ value of agent $a_i$'s domain which is calculated by the neighbor $a_j$, $x_{j,l}$ indicates that $x_j$ is assigned the $l^{th}$ value of $a_j$'s domain, $D_j$, $C$ is the cost for the function which is an element of all the constraint function set $F_j$ between agent $a_i$ and $a_j$, $\widetilde{x_j}$ is the current partial assignment sent from $a_i$ to $a_j$ that contains the known assigned values of the neighbors of $a_i$, $x_{i,k}$ indicates that $x_i$ is assigned the $k^{th}$ value of agent $a_i$'s domain $D_i$. Agent $a_j$ calculates $\zeta_{j,k}$ for all the values of $k \in D_i$ and the resulting cost map $\zeta_j$ = \{$\zeta_{j,1}$, $\zeta_{j,2}$, . . . . , $\zeta_{j,|D_i|}$\} is sent to the inquiring agent $a_i$. Then, $a_i$ finds the value of its variable $x_i$ from (Equation~\ref{eq:ea2}). Here, $\delta$ is the minimum aggregated cost received from the neighbors for each $k \in D_i$, $\rho$ is a set of values from agent $a_i$'s domain for which the cost is minimum and $\zeta_{j,k}$ is the received cost messages from its neighbors.

\begin{equation}\label{eq:21}
\zeta_{j,k} = \min_{x_{j,l} \in D_j} \sum_{C \in F_j} C(\widetilde{x_j} \cap x_{i,k} \cap x_{j,l})  \\
\end{equation}
\begin{equation}\label{eq:ea2}
\delta = min \sum_{j = 1}^{|\mathcal{N}_i|} \zeta_{j,k}; \hspace{0.3cm} \rho = \{k: \sum_{j = 1}^{|\mathcal{N}_i|} \zeta_{j,k} = \delta\} 
\end{equation}

Notably, for more than one value in $a_i$'s domain in $\rho$, the \textit{unique-first} approach is followed to determine whether the current solution is accepted or not. In this approach, $|\rho|$ is compared with a bound $\beta$. The initial value of $\beta$ is set to 1. This means that the value is acceptable if it is a unique local optimum. If $|\rho| > \beta$, agent $a_i$ goes into HOLD state and waits for more information. Otherwise, a value is selected randomly from $\rho$ and is assigned to its controlled variable. After assigning a value to $x_i$, every agent $a_j \in \mathcal{N}_i$ updates its current partial assignment and repeats the algorithm. If the value assignment is not  possible for all the agents, $\beta$ is increased by 1, and the algorithm is repeated. This approach prevents the agents from assigning a value prematurely to their variables.

\subsection{Challenges}
We need to address the following challenges to develop an F-DCOP algorithm that adapts CoCoA.

\begin{itemize}
    \item \textbf{Infinite Domain:} For F-DCOPs, the domain is an infinite number of values within a range. In effect, an agent needs to assign a value to its variables from an infinite number of points. Thus, an F-DCOP solver requires an extensive amount of time and memory to converge.
    \item \textbf{Discretization:} F-DCOP solvers need to discretize the continuous state space to operate. The choice of discrete points can be random; however, setting up the number of discrete points is critical. The quality of solutions found by an F-DCOP algorithm increases with the increasing number of points.
    \item \textbf{Initializing Parameters:} If the cost functions are not convex, initializing the parameters in continuous non-linear optimization methods is significant. Because, even with infinite computing power and time, the gradient approach can still stuck with local minimum or\:saddle\:point.
\end{itemize}
In the following section, we devise a novel method to apply CoCoA in F-DCOPs.
\section{Continuous Cooperative Constraint Approximation (C-CoCoA)}
\label{sec:3}

To address the challenges discussed in the previous section, we propose C-CoCoA, a non-exact algorithm  that uses Cooperative Constraint Approximation (CoCoA) as the underlying algorithmic framework. To be precise, we combine the discrete CoCoA algorithm and the continuous non-linear optimization technique. C-CoCoA is also a non-iterative algorithm like CoCoA in the sense that each agent can only assign its value once and once assigned, it cannot change its value.
\subsection{C-CoCoA: Algorithm Description}
C-CoCoA (i.e. Algorithm~\ref{algo:C-CoCoA}) defines $\mathcal{N}_i$ as the set of direct neighbors of the agent $a_i$. We assume that, an agent $a_i$ communicates only with those agents whose variables affect $a_i$'s cost function. In other words, $a_i$ communicates only with $a_j \in \mathcal{N}_i$. This ensures a low communication overhead as well as a fully decentralized solution. For this reason, the total cost of an individual agent $a_i$ only depends on $|\mathcal{N}_i|$ rather than the size of the constraint graph. We also assume that each agent knows its neighbors' discretized domain and the nodes of the constraint graph are reachable from any other node. 
\begin{algorithm}[t]
\small
\caption{The C-CoCoA Algorithm}\label{algo:C-CoCoA}
\DontPrintSemicolon
\SetKwFunction{FUpdateState}{UpdateState}
\SetKwFunction{FInquiryMSG}{InquiryMSG}
\SetKwFunction{FIdleActiveNeighbors}{IdleActiveNeighbors}
\SetKwFunction{FGradientDescent}{GradientDescent}
\SetKwFunction{FSetValue}{SetValue}
\SetKwFunction{FInit}{Init}
\SetKwInOut{Input}{input}
\SetKwInOut{Output}{output}
\SetKwInOut{Require}{require}
\Input{A constraint graph $G$, set of agents $A$, set of variables $X$, number of discrete points $k$, $\beta$}
\Output{Near-optimal assignment of the variables that minimizes the overall cost}

  \BlankLine
  Discretize the domain of the variables into $k$ points, $x_i(1)$, $x_i(2)$, . . . , $x_i(k)$\;

  $STATE \leftarrow$ ACTIVE, HOLD or DONE\;
  $CPA \leftarrow$ current partial assignment\;
  $\psi \leftarrow$ a set of agents with $STATE$ := DONE\;

  	\For{each agent $a_i \in A$}{
  		    $STATE_{a_i} \leftarrow$ IDLE\;
  		    $CPA_{a_i} \leftarrow$ \{ \} \;
  	}
  	$\psi \leftarrow$ \{ \}, $\beta$ = 1 \;
  	\BlankLine
  	randomly select any agent $a_i$ from the set $A-\psi$\;
  	$STATE_{a_i} \leftarrow$ ACTIVE\;
  	\For{each agent $a_j \in \mathcal{N}_i$}{
  		\FUpdateState{$i$, $j$, $ACTIVE$}\;
  		$\zeta_j$ $\leftarrow$ \FInquiryMSG{$i$, $j$, $CPA_{a_i}$}\;
  	}
  	calculate $\rho$ using Equation~\ref{eq:ea2}\;
  	$\chi \leftarrow$ values of $x_j$ that results $\zeta_j$ $\hspace*{1cm}$\;
  	
	 \If{$|\rho| \leq \beta$ {\bf or} \FIdleActiveNeighbors{$i$} $=$ 0}{
		$\Theta_{x_i}$ $\leftarrow$ randomly select a value from $\rho$\;
		$\chi$ $\leftarrow$ $\chi \cup \Theta_{x_i}$\;
  		calculate $F^{a_i}_{\mathcal{N}_i}$ using Equation~\ref{eq:4}\;
  		$x^{a_i}_{\mathcal{N}_i} \leftarrow$ set of related variables with $F^{a_i}_{\mathcal{N}_i}$\;
  		\For{each variable $x \in x^{a_i}_{\mathcal{N}_i}$}{
  		    initialize $x$ with the corresponding value from $\chi$\;
  		}
  		\While{the terminating condition is not met}{
  		    $\forall x \in x^{a_i}_{\mathcal{N}_i}$ update $v_x$ using Equation~\ref{eq:5}\;
  		}
  		$x_i$ $\leftarrow$ $v_{x_i}$\;
  		$STATE_{a_i} \leftarrow$ DONE\;
  		$\psi \leftarrow \psi \cup a_i$\;
  		\For{each agent $a_j \in \mathcal{N}_i$}{
  			\FUpdateState{$i$, $j$, $DONE$}\;
  			\FSetValue{$i$, $CPA_{a_i}$}\;
  		}			 
	 }
	 \Else{
	 	$STATE_{a_i} \leftarrow$ HOLD\;
	 	\For{each agent $a_j \in \mathcal{N}_i$}{
  			\FUpdateState{$i$, $j$, $HOLD$}\;
  		}
	 }
\end{algorithm}
\setlength{\textfloatsep}{0pt}

The C-CoCoA algorithm uses the same message passing technique as described in Section~\ref{sec:cocoa} for the discrete CoCoA, using the current discretizations of the domain of each variable $x_i$. However, as the cost functions are not in the tabular form, each agent calculates the cost by evaluating $C = f_i(x^i)$, where $x^i$ is the set of variables related to $f_i$.

The key difference between the C-CoCoA and discrete CoCoA is that, in C-CoCoA each agent $a_i$ calculates the cost by considering its domain discretizations $x_i(1)$, $x_i(2)$,..., $x_i(k)$ (Algorithm~\ref{algo:C-CoCoA}: Line 1) instead of the actual continuous domain, where $k$ is the total number of random discrete points taken from $D_i$. We select the discrete 
points randomly because, as aforementioned, we use the non-linear optimization technique to adjust these random discrete points later. For the example of Figure~\ref{fdcopex}, for simplicity, let us assume that $k$ = 2. So, we discretize the domains of $x_0, x_1, x_2$ and $x_3$ into 2 random discrete points ($x_0$: [1, 2], $x_1$: [3, 4], $x_2$: [7, 8] and $x_3$: [5, 9]) from the domain range [-20, 20].

\begin{procedure}[h]
\small
\caption{InquiryMSG()}\label{proc: inq}

\DontPrintSemicolon
\SetKwFunction{FInquiryMSG}{InquiryMSG}
\SetKwProg{Fn}{Function}{:}{}
  	\Fn{\FInquiryMSG{$i$, $j$, $CPA_{a_i}$}}{
  	    $\zeta \leftarrow$ \{ \}\;
  		\For{all $x_{i,k}$ $in$ $D_i$}{
  		    $cost \leftarrow$ \{ \}\;
  		    \If{$a_j \in CPA_{a_i}$}{
  		        $D_j \leftarrow$ value of $a_j$\;
  		    }
  			\For{all $x_{j,l}$ $in$ $D_j$}{
  				calculate $\zeta_{j,k}$ using Equation~\ref{eq:21}\;
  				$cost \leftarrow cost \cup \zeta_{j,k}$
  			}
  			$C \leftarrow \min(cost), v_j \leftarrow \argmin_j(C)$\;
  			$\zeta \leftarrow \zeta \cup \{v_j:C\}$\;
  		}
  		\Return $\zeta$
  	}
\end{procedure}

\begin{figure*}
\centering

  \begin{tikzpicture}
        [
        roundnode/.style={circle, draw=black, fill=white!5, very thick, minimum size=5mm, scale=0.8},
        ]
        \node[roundnode, fill={rgb:black,0.2;white,2}]    at(-13.5,2.5)  (x0)                                             {$a_0$};
        \node at (-13.5,3) {[1,2]};
        \node[roundnode, fill=yellow]    at(-15,2.5)  (x3)                              {$a_3$};
        \node at (-15,3) {[5,9]};
        \node[roundnode, fill=yellow]    at(-12,3)  (x1)                              {$a_1$};
        \node at (-12,3.65) {[3,4]};
        \node[roundnode, fill=yellow]    at(-12,1.5)  (x2)                              {$a_2$};
        \node at (-13,1.5) {[7,8]};
         
        \draw[->, line width=0.5mm, blue] (x0) -- (x3);
        \draw[->, line width=0.5mm, blue] (x0) -- (x1);
        \draw[->, line width=0.5mm, blue] (x0) -- (x2);
        \draw (x1) -- (x2);
        \node  at (-13.5,1)
        {
            (a) Inquiry messages from $a_0$
        };
        
        [
        roundnode/.style={circle, draw=black, fill=white!5, very thick, minimum size=7mm},
        ]
        \node[roundnode, fill={rgb:black,0.2;white,2}]    at(-9,2.5)  (x0)                                             {$a_0$};
        \node[roundnode, fill=yellow]    at(-10.5,2.5)  (x3)                              {$a_3$};
        \node[roundnode, fill=yellow]    at(-7.5,3)  (x1)                              {$a_1$};
        \node[roundnode, fill=yellow]    at(-7.5,1.5)  (x2)                              {$a_2$};

        \draw[->, line width=0.5mm, red] (x3) -- (x0);
        \draw[->, line width=0.5mm, red] (x1) -- (x0);
        \draw[->, line width=0.5mm, red] (x2) -- (x0);
        \draw (x1) -- (x2);
        \node  at (-9,1)
        {
            (b) Cost messages to $a_0$
        };
        
        [
        roundnode/.style={circle, draw=black, fill=white!5, very thick, minimum size=7mm},
        ]
        \node[roundnode, fill=yellow]    at(-4.5,2.5)  (x0)                                             {$a_0$};
        \node[roundnode]    at(-6,2.5)  (x3)                              {$a_3$};
        \node[roundnode, fill={rgb:black,0.2;white,2}]    at(-3,3)  (x1)                              {$a_1$};
        \node[roundnode, fill=yellow]    at(-3,1.5)  (x2)                              {$a_2$};

        \draw[->, line width=0.5mm, blue] (x1) -- (x0);
        \draw[->, line width=0.5mm, blue] (x1) -- (x2);
        \draw (x0) -- (x2);
        \draw (x0) -- (x3);
        \node  at (-4.5,1)
        {
            (c) Inquiry messages from $a_1$
        };
        
        [
        roundnode/.style={circle, draw=black, fill=white!5, very thick, minimum size=7mm},
        ]
        \node[roundnode, fill=yellow]    at(0,2.5)  (x0)                                             {$a_0$};
        \node[roundnode]    at(-1.5,2.5)  (x3)                              {$a_3$};
        \node[roundnode, fill={rgb:black,0.2;white,2}]    at(1.5,3)  (x1)                              {$a_1$};
        \node[roundnode, fill=yellow]    at(1.5,1.5)  (x2)                              {$a_2$};

        \draw[->, line width=0.5mm, red] (x0) -- (x1);
        \draw[->, line width=0.5mm, red] (x2) -- (x1);
        \draw (x2) -- (x0);
        \draw (x0) -- (x3);
        \node  at (0,1)
        {
            (d) Cost messages to $a_1$
        };
\end{tikzpicture}
\caption{Message passing process of the C-CoCoA algorithm to solve the F-DCOP shown in Figure~\ref{fdcopex}.}
\label{fig:2}
\end{figure*}
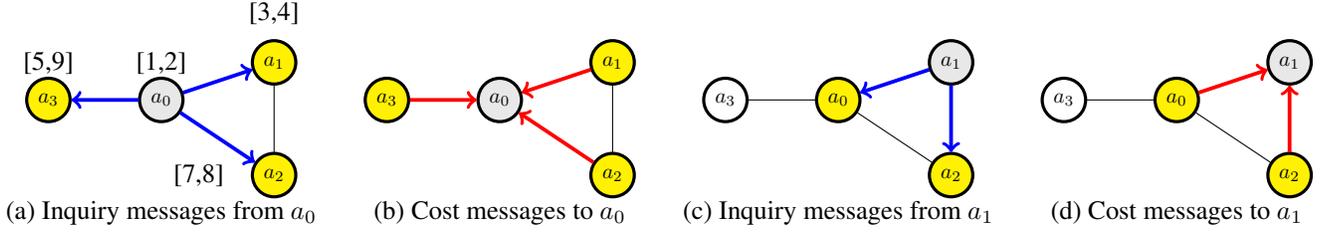

The states of the agents are defined as IDLE, ACTIVE, HOLD and DONE \cite{van2017cocoa}. The current partial assignment (CPA) denotes the known assigned values of the neighbors of $a_i$. We define a set $\psi$ that contains the set of agents who finish their variable assignments. Therefore, $A-\psi$ is the set of unassigned agents (value assignment to their variables is not finished). Then in the initialization step, each agent $a_i$ initializes its state to IDLE, the current partial assignment with an empty assignment and $\psi$ with an empty set. (Algorithm~\ref{algo:C-CoCoA}: Line 5-8). After this step, similar to the discrete CoCoA algorithm, our algorithm activates an agent $a_i$ randomly, because, starting with any agent yields the same result. Agent $a_i$ activates each agent $a_j \in \mathcal{N}_i$ (Procedure~\ref{proc: upd}) and sends an inquiry message (Procedure~\ref{proc: inq}) to each of the $a_j$ (Algorithm~\ref{algo:C-CoCoA}: Line 9-13). We define $\zeta_j = \{\zeta_{j,x_i(1)}, \zeta_{j,x_i(2)},..., \zeta_{j,x_i(k)}\}$ as the overall cost map that contains the minimum cost for each of the discrete points of $a_i$'s domain and is calculated by the agent $a_j$. We define each element of the cost map as $\zeta_{j,x_i(k)} = \{v_j:C\}$, where, $C$ is the minimum cost and $v_j$ = $\argmin_j{(C)}$ denotes the value of $a_j$'s domain that gives the minimum cost $C$. The agent $a_i$ then calculates $\rho$ using the Equation~\ref{eq:ea2}. $\rho$ contains the values of $x_i$ that is near-optimal within the discretized points $x_i(1)$, $x_i(2)$,..., $x_i(k)$ of the agent $a_i$'s domain. $a_i$ also stores the values of $v_j \in \zeta_{j,x_i(k)}$ in a set $\chi$ (Algorithm~\ref{algo:C-CoCoA}: Line 14-15). For the example of Figure~\ref{fdcopex}, we assume that the agent $a_0$ is selected randomly. $a_0$ then activates its neighbors $a_1$, $a_2$, $a_3$ and sends an inquiry message to all of them. Upon receiving the inquiry message, $a_1, a_2$ and $a_3$ calculate the cost map $\zeta_1$ = \{3: 13, 3: 10\}, $\zeta_2$ = \{7: 154, 7: 161\}, $\zeta_3$ = \{5: 30, 5: 35\} respectively (Figure~\ref{fig:2}) and send these cost maps to the inquiring agent $a_0$. After receiving the cost map, $a_0$ calculates $\rho$ by using the Equation~\ref{eq:ea2} and for this example $a_0$ assigns $\rho = \{1\}$ and $\chi$ = \{$x_1$ = 3, $x_2$ = 7, $x_3$ = 5\}. We describe this example elaborately in the Figure~\ref{fig:2}.

\begin{procedure}[h]
\small
\caption{UpdateState()}\label{proc: upd}
\DontPrintSemicolon
\SetKwFunction{FUpdateState}{UpdateState}
\SetKwProg{Fn}{Function}{:}{}
  	\Fn{\FUpdateState{$i$, $j$, $S$}}{
        $a_j$ sets $STATE_{a_i}$ $\leftarrow$ $S$\;
        \If{$S$ = HOLD {\bf and} $STATE_{a_j}$ = HOLD {\bf and} \FIdleActiveNeighbors{$i$} = 0}{
        	$\beta++$\;
        	Goto Line 9, Algorithm~\ref{algo:C-CoCoA}\;
        }
        \If{$S$ = DONE {\bf and} $STATE_{a_j}$ = HOLD}{
            Goto Line 9, Algorithm~\ref{algo:C-CoCoA}\;
        }
  	}
\end{procedure}

Similar to the discrete CoCoA algorithm, more than one value in $a_i$'s domain can achieve the minimum cost \cite{van2017cocoa}, that is $|\rho| > 1$. In this case, we follow a \textit{unique-first} approach which is described in the CoCoA algorithm (Section~\ref{sec:cocoa}). Algorithm~\ref{algo:C-CoCoA}: Line 31-34, describes the case when $|\rho| > \beta$. In this case, $a_i$ goes into HOLD state and waits until another agent has completed its assignment and repeats the Algorithm~\ref{algo:C-CoCoA} from Line 9. Otherwise (when $|\rho| \leq \beta$), a value is selected randomly from the set $\rho$. We assign this value to $\Theta_{x_i}$ and add this $\Theta_{x_i}$ to the set $\chi$ (Algorithm~\ref{algo:C-CoCoA}: Line 17-18). This assignment is near-optimal within the discretized domain. In order to find the best solution within the actual domain $D_i$, we use a non-linear optimization technique. We choose gradient-based optimization approach because we can implement it in a decentralized way using only local information. Now, for employing the gradient-based non-linear optimization, agent $a_i$ calculates the local objective function $F^{a_i}_{\mathcal{N}_i}$ (Algorithm~\ref{algo:C-CoCoA}: Line 19) by using the following equation: 
\begin{equation}
    F^{a_i}_{\mathcal{N}_i} = \sum_{a_j \in \mathcal{N}_i} f{(a_i, a_j)} \label{eq:4} \\
\end{equation}
where, $f{(a_i, a_j)}$ is the cost function that is related to agent $a_i$ and its direct neighbor $a_j \in \mathcal{N}_i$. For the example of Figure~\ref{fdcopex}, agent $a_0$ assigns $\Theta_{x_0}$ = 1 from $\rho$ and appends this value with the set $\chi$. Hence, the set $\chi$ = \{$x_0 = 1, x_1 = 3, x_2 = 7, x_3 = 5$\}. Thereafter, the agent $a_0$ calculates the local objective function $F^{a_i}_{\mathcal{N}_i}$ = $x_0^2 - 2x_0x_1 + 2x_1^2 + x_0x_2 + 3x_2^2 + x_0x_3 + x_3^2$.
\setlength{\textfloatsep}{2mm}
\setlength{\floatsep}{2mm}

After that, the agent $a_i$ performs gradient-based approach for optimizing its local objective function $F^{a_i}_{\mathcal{N}_i}(x^{a_i}_{\mathcal{N}_i})$ where, $x^{a_i}_{\mathcal{N}_i}$ is the set of all the related variables with $F^{a_i}_{\mathcal{N}_i}$ (Algorithm~\ref{algo:C-CoCoA}: Line 20). Agent $a_i$ assigns every variable $x \in x^{a_i}_{\mathcal{N}_i}$ with the corresponding value from the set $\chi$ as the initial values in the gradient-based optimization method (Algorithm~\ref{algo:C-CoCoA}: Line 21-22). Specifically, the agent $a_i$ minimizes the local objective function $F^{a_i}_{\mathcal{N}_i}$ and updates the value $v_x$ of each variable $x \in x^{a_i}_{\mathcal{N}_i}$ according to the following equation:
\begin{equation}
    v_x(t) = v_x(t-1) - \alpha \frac{\partial F^{a_i}_{\mathcal{N}_i}}{\partial x^{a_i}_{\mathcal{N}_i}} \bigg|_{\argmin_{x_i} F^{a_i}_{\mathcal{N}_i}(x^{a_i}_{\mathcal{N}_i} = v_x)}^{v_x} \label{eq:5} \\
\end{equation}
where $\alpha$ is the \textit{learning rate} of the algorithm (Algorithm~\ref{algo:C-CoCoA}: Line 23-24). For the example of Figure~\ref{fdcopex}, $x^{a_0}_{\mathcal{N}_0}$ = \{$x_0, x_1, x_2, x_3$\}. Agent $a_0$ initializes all the variables in $x^{a_0}_{\mathcal{N}_0}$ from the set $\rho$ in the gradient-based optimization. In this example, the initial values are set as $x_0 = 1, x_1 = 3, x_2 = 7, x_3 = 5$. Then the agent $a_0$ starts updating the values of the variables $x^{a_0}_{\mathcal{N}_0}$ by using the Equation~\ref{eq:5}.

\begin{procedure}[h]
\small
\caption{IdleActiveNeighbors()}\label{proc: ian}
\DontPrintSemicolon
\SetKwFunction{FIdleActiveNeighbors}{IdleActiveNeighbors}
\SetKwProg{Fn}{Function}{:}{}
  	\Fn{\FIdleActiveNeighbors{$i$}}{
  		\For{each agent $a_j \in \mathcal{N}_i$}{
  			\If{$STATE_{a_j}$ = IDLE {\bf or} ACTIVE}{
  				$count++$\;
  			}
  		}
  		\Return $count$
  	}
\end{procedure}
\setlength{\textfloatsep}{0pt}

The agent continues this update process until it converges or a maximum number of iterations is reached. After termination, the current value of $v_x$ is actually the approximate optimal assignment for the variable $x_i$ (Algorithm~\ref{algo:C-CoCoA}: Line 25). Then the agent $a_i$ updates its state to DONE, updates the set $\psi$ and communicates to its neighbors $a_j \in \mathcal{N}_i$ in a SetValue message (Algorithm~\ref{algo:C-CoCoA}: Line 26-30). By receiving this message, each neighbor $a_j$ updates its CPA with the value of $x_i$ and repeats the Algorithm~\ref{algo:C-CoCoA} from Line 9 for the unassigned agents ($a_i \in A-\psi$). When the set $A-\psi$ is empty (all the agents finish their variable assignment), the algorithm terminates. For our example, after 100 iterations, the final assignments for these variables are \{$x_0 = -0.572$\}. Then the agent $a_0$ assigns $x_0$ with this final value. Agent $a_0$'s state is marked as done, $a_0$ is added to the set $\psi$ and $a_0$ sends a SetValue message to all the neighbors of $a_0$ ($a_1, a_2, a_3$). The neighbors update their CPA with $x_0$ and repeats the Algorithm~\ref{algo:C-CoCoA} from Line 9 for the unassigned agents. Note that, each agent can only assign its value once and once assigned it cannot change its value. To be precise, each agent updates its value locally with gradient descent and sends the setValue() message only once to a neighbor and thus C-CoCoA is a non-iterative approach.

\section{Worked Example}

This section describes a complete example of our algorithm C-CoCoA. We use the F-DCOP shown in Figure~\ref{fdcopex} as the example problem and show the result that is obtained by the C-CoCoA algorithm. In this example, we assume that the variable $x_i$ is controlled by the agent $a_i$ and $x_0$: [1, 2], $x_1$: [3, 4], $x_2$: [7, 8], $x_3$: [5, 9] are the 2 random discrete points taken from the actual domain [-20, 20]. We also assume that, $\zeta_{j,k}$ is the cost for the $k^{th}$ value of agent $a_i$'s domain which is calculated by neighbor $a_j$ and $\zeta_j$ is the overall cost map that is calculated by the neighbor $a_j$. We use the arrows between the nodes of the constraint graph to indicate the direction of the corresponding messages. Figure~\ref{fig:2} shows the message passing process and the results are as follows:

$\bullet$ C-CoCoA Algorithm starts by randomly selecting an agent $a_0$. $a_0$ sends inquiry message to $a_1, a_2$ and $a_3$, blue arrows represent the inquiry messages, grey node represents the inquiring agent. $a_1, a_2$ and $a_3$ calculates the cost map, $\zeta$, yellow nodes represent the active neighbors (Figure~\ref{fig:2}(a)). 

$\bullet$ Agent $a_1$ calculates $\zeta_{3, 1}$ = 13, $\zeta_{4, 1}$ = 25, therefore, appends [3: 13] with $\zeta_1$. The agent also calculates $\zeta_{3, 2}$ = 10, $\zeta_{4, 2}$ = 20, therefore, appends [3: 10] with $\zeta_1$. $a_1$ sends the final cost map $\zeta_1$ = [3: 13, 3: 10] to $a_0$. Agent $a_2$ calculates $\zeta_{7, 1}$ = 154, $\zeta_{8, 1}$ = 200, therefore, appends [7: 154] with $\zeta_2$. The agent also calculates $\zeta_{7, 2}$ = 161, $\zeta_{8, 2}$ = 208, therefore, appends [7: 161] with $\zeta_2$. $a_2$ sends the final $\zeta_2$ = [7: 154, 7: 161] to $a_0$. Agent $a_3$ calculates $\zeta_{5, 1}$ = 30, $\zeta_{9, 1}$ = 90, therefore, appends [5: 30] with $\zeta_3$. The agent also calculates $\zeta_{5, 2}$ = 35, $\zeta_{9, 2}$ = 99, therefore, appends [5: 35] with $\zeta_3$. $a_3$ sends the final $\zeta_3$ = [5: 30, 5: 35] to $a_0$.

$\bullet$ Agent $a_0$ receives cost maps from $a_1, a_2$ and $a_3$, red arrows represent the cost messages (Figure~\ref{fig:2}(b)). Agent $a_0$ calculates $\rho$ using Equation~\ref{eq:ea2}.
           For the discretized domain value 1, $a_0$ calculates cost = 13 + 154 + 30 = 197.
           For the discretized domain value 2, $a_0$ calculates cost = 10 + 161 + 35 = 206. Therefore, $\rho$ = 1 and $\chi$ = \{$x_0=1, x_1=3, x_2=7, x_3=5$\}. After 100 iterations of the gradient-based approach, we get, $x_0$ = -0.572.

$\bullet$ After the completion of $a_0$, algorithm selects the agent $a_1$. $a_1$ sends inquiry message to $a_0$ and $a_2$, blue arrows represent the inquiry messages (Figure~\ref{fig:2}(c)). Agents $a_0$ and $a_2$ calculate the cost maps.

$\bullet$ Agent $a_0$ calculates $\zeta_{-0.572, 3}$ = 21.756, therefore, appends [-0.572: 21.756] with $\zeta_0$. The agent also calculates $\zeta_{-0.572, 4}$ = 36.899, therefore, appends [-0.572: 36.899] with $\zeta_0$. $a_0$ sends the final $\zeta_0$ = [-0.572: 21.756, -0.572: 36.899] to $a_1$. Agent $a_2$ calculates $\zeta_{7, 3}$ = 86, $\zeta_{8, 3}$ = 113, therefore, appends [7: 86] with $\zeta_2$. The agent also calculates $\zeta_{7, 4}$ = 86, $\zeta_{8, 4}$ = 112, therefore, appends [7: 86] with $\zeta_2$. $a_2$ sends the final $\zeta_2$ = [7: 86, 7: 86] to $a_1$.

$\bullet$ Agent $a_1$ receives cost maps from $a_0$ and $a_2$, red arrows represent the cost messages (Figure~\ref{fig:2}(d)). Agent $a_1$ calculates $\rho$ using Equation~\ref{eq:ea2}.
          For the discretized domain value 3, $a_0$ calculates cost = 21.756 + 86 = 107.756.
          For the discretized domain value 4, $a_0$ calculates cost = 36.899 + 86 = 122.899. Therefore, $\rho$ = 3 and $\chi$ = \{$x_0=-0.572, x_1=3, x_2=7$\}. After 100 iterations of the gradient-based approach, we get, $x_1$ = -0.122.
          
$\bullet$ Agents $a_2$ and $a_3$ calculate the values for their variables by repeating the algorithm. We get $x_2$ = 0.124 and $x_3$ = 0.911 when it terminates.\:Hence, the near-optimal assignment is, $\mathbf{X^*}$ = \{$\mathbf{x_0}$ = \textbf{-0.572}, $\mathbf{x_1}$ = \textbf{-0.122}, $\mathbf{x_2}$ = \textbf{0.124} and $\mathbf{x_3}$ = \textbf{0.911}\}

\begin{figure*}
  \centering
  \begin{subfigure}[b]{0.3\linewidth}
    \centering\includegraphics[height=4cm, width=5.9cm]{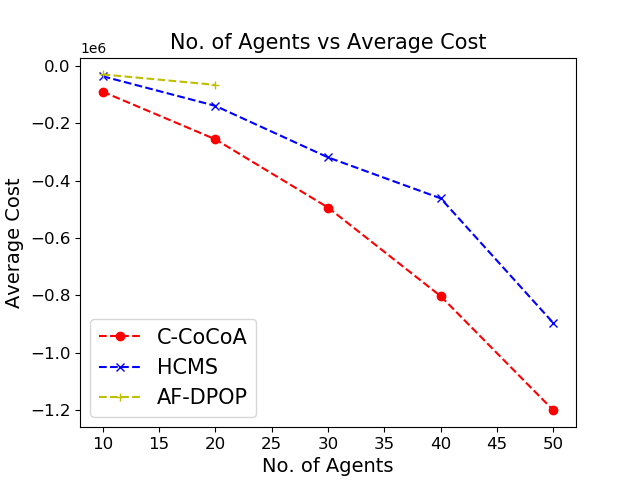}
    \caption{\label{fig:sparse} Sparse graph}
  \end{subfigure}%
  \begin{subfigure}[b]{0.3\linewidth}
    \centering\includegraphics[height=4cm, width=5.9cm]{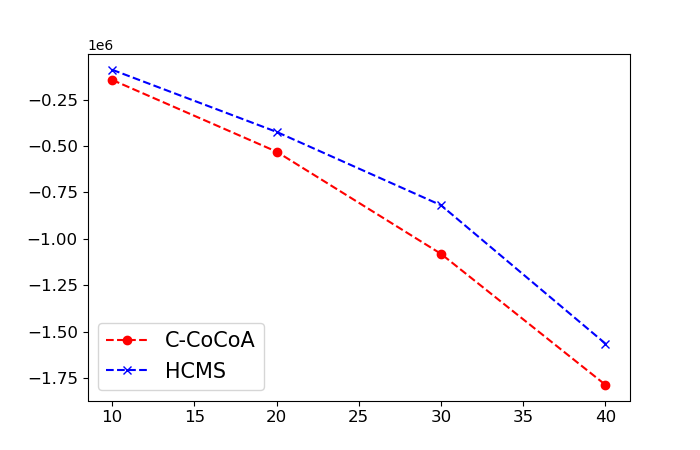}
    \caption{\label{fig:dense} Dense graph}
  \end{subfigure}
  \begin{subfigure}[b]{0.3\linewidth}
    \centering\includegraphics[height=4cm, width=5.9cm]{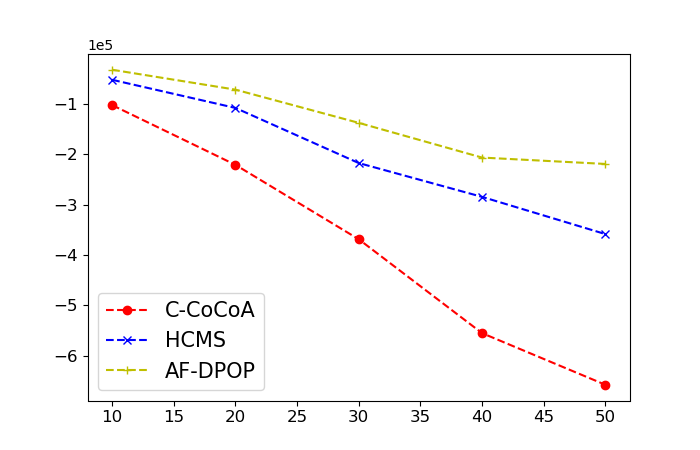}
    \caption{\label{fig:scalefree} Scale-free graph}
  \end{subfigure}
  \caption{Solution cost comparison of C-CoCoA and the competing algorithms varying the number of agents.}
\end{figure*}
	\setlength{\textfloatsep}{0mm}
	\setlength{\floatsep}{0mm}

\section{Complexity Analysis}

In C-CoCoA, we define the total number of agents $|A| = n$ and $\mathcal{N}_i$ is the set of direct neighbors of the agent $a_i$. After the activation of an agent $a_i$, it sends $2 * |\mathcal{N}_i|$ messages (UpdateState and InquiryMSG) to its neighbors as well as $a_i$ receives $|\mathcal{N}_i|$ messages from its neighbors against the reply of the inquiry messages. Therefore, at this stage (Algorithm~\ref{algo:C-CoCoA}: Line 11-13), the number of messages is $3 * |\mathcal{N}_i|$. Then, after a successful assignment to its variable, the agent $a_i$ sends $2 * |\mathcal{N}_i|$ messages (UpdateState and SetValue) to its neighbors (Algorithm~\ref{algo:C-CoCoA}: Line 28-30). As a result, the number of messages transmitted so far is $5 * |\mathcal{N}_i|$. However, an agent sends additional $|\mathcal{N}_i|$ messages each time it enters into the HOLD state (Algorithm~\ref{algo:C-CoCoA}: Line 33-34).
Although an agent may never enter into the HOLD state, in the worst case, it may enter into the HOLD state $k$ times at most, where, $k$ is the total number of discrete points taken from the agents' domain. For this reason, $5 * |\mathcal{N}_i| + H * |\mathcal{N}_i|$ is the total number of messages an agent sends and receives, where, $H = 0,1,...,k$ defines the number of times an agent enters into the HOLD state. In the worst case, the graph is complete where, $|\mathcal{N}_i| = n-1 \approx n$ and $H = k$. Therefore, the total number of messages sent or received by an agent $a_i$ is $O(5n + kn)$ in the worst case.

The size of each UpdateState message is constant and in each of the InquiryMSG and SetValue message, the agent $a_i$ sends the $CPA_{a_i}$ that contains the set of known assigned values of all the neighbors $a_j \in \mathcal{N}_i$. Hence, the size of each InquiryMSG and SetValue message is $|\mathcal{N}_i|$. $a_i$ sends total $|\mathcal{N}_i|$ InquiryMSG and SetValue messages to its neighbors. So, the summation of message size complexity of InquiryMSG and SetValue messages is $|\mathcal{N}_i|^2 + |\mathcal{N}_i|^2 = 2|\mathcal{N}_i|^2$. When the neighboring agents send inquiry message to $a_i$, it sends a reply message of size $k$ as well that contains the cost map $\zeta$. Therefore, $a_i$ sends $|\mathcal{N}_i|$ reply messages of size $k$ to the neighbors. Hence, the total message size for an agent $a_i$ is $O(2|\mathcal{N}_i|^2 + k|\mathcal{N}_i|) \approx O(2n^2 + kn)$ in the worst case in C-CoCoA.

After the initialization and the transmission of UpdateState and InquiryMSG (Algorithm~\ref{algo:C-CoCoA}: Line 5-13), the computational complexity of an agent is $|\mathcal{N}_i| + |\mathcal{N}_i| * k^2$ ($k^2$ is the complexity of calculating an InquiryMSG). In the gradient-based optimization, an agent needs $|x^{a_i}_{\mathcal{N}_i}| + b * |x^{a_i}_{\mathcal{N}_i}|$ computational complexity, where, $b$ is the number of times an agent updates the values of the variables (Algorithm~\ref{algo:C-CoCoA}: Line 21-24). After a successful assignment or each of the unsuccessful attempt (HOLD state) to assign a value, an agent again iterates over the set of its neighbors (Algorithm~\ref{algo:C-CoCoA}: Line 28-34). This step adds $|\mathcal{N}_i| + H * |\mathcal{N}_i|$ complexity. After adding all these, the overall computational complexity is $O(n+n*k^2+n+n*b+n+H*n) \approx O(n(k^2+b))$; where, in the worst case $|\mathcal{N}_i| \approx n$, $|x^{a_i}_{\mathcal{N}_i}| \approx n$ and $H = k$.

\section{Experimental Results}\label{sec:5}
In this section, we empirically evaluate the performance of C-CoCoA with HCMS and AF-DPOP. The performance metrics are solution quality, time, and number of messages. Two types of graphs are used for comparison, namely, \textit{Random Graphs} and \textit{Random Trees}. Although \cite{hoang2019new} proposed three versions of Functional DPOP, we only compare with AF-DPOP in this paper. The reason is AF-DPOP is reported to provide the best solution among the approximate algorithms proposed in their work. However, CMS is not used in the benchmark since it uses only piecewise linear functions which are not applicable for most of the real-world problems. For all the experiments, binary quadratic functions are used which are of the form $ax^2 + bxy + cy^2$. However, it is worth mentioning that although we choose binary quadratic functions for evaluation, C-CoCoA is broadly applicable to other classes of problems. We choose coefficients of the cost functions $(a, b, c)$ randomly between $[-5, 5]$ and set the domains of each agent to $[-50, 50]$. The averages are taken over 50 randomly generated problems. The experiments are carried out on a machine with an Intel core i5-6500 cpu, 3.2 GHz processor and 8 GB RAM.\par

\begin{figure}[t]
\centering
  \includegraphics[height=4cm, width=7.9cm]{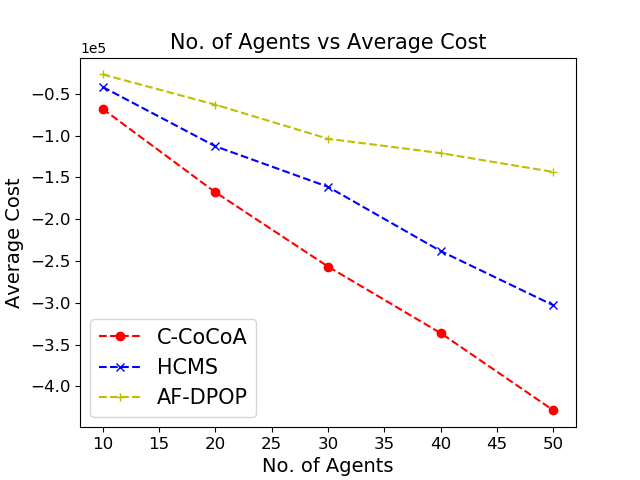}
  \caption{Solution Cost Comparison of C-CoCoA and the competing algorithms varying the number of agents (random trees)}
  \label{fig:tree}
\end{figure}

\textbf{Random Graphs:} We use three different settings for random graphs - sparse, dense and scale-free. For all the algorithms, we choose the number of discrete points to be 3. However, we compare the performance of C-CoCoA varying the number of discrete points later in this section. For C-CoCoA, we set the maximum number of iterations for Equation \ref{eq:5} to be 100 and $\alpha=0.01$ (which is the best result found on the empirical evaluation). Moreover, we stop HCMS after 100 iterations in Figures \ref{fig:sparse}, \ref{fig:dense}, \ref{fig:scalefree} and \ref{fig:tree}. Note that, although AF-DPOP requires fewer messages than HCMS, we do not limit the number of messages for HCMS since AF-DPOP requires much more computation to calculate one message. The detailed analysis of computation, time, and number of messages for each of the algorithms are given in Table \ref{tab:all}. Figure \ref{fig:sparse} shows the comparison of average costs on Erd{\H{o}}s-R{\'e}nyi topology \cite{erdHos1960evolution} with sparse settings (edge probability 0.2) varying the number of agents. This figure shows that C-CoCoA performs better than both HCMS and AF-DPOP on average. For $no. \ of \ agents \geq 20$, AF-DPOP runs out of memory. Thus, we omit the result of AF-DPOP for $no. \ of \ agents \geq 20$.\par

We choose dense graphs as our second random graph settings. Figure \ref{fig:dense} shows the cost comparison between the C-CoCoA and HCMS on Erd{\H{o}}s-R{\'e}nyi topology with dense settings (edge probability 0.6). C-CoCoA shows comparatively better performance than HCMS. Note that, AF-DPOP is not used in the dense graph setting due to the huge computation overhead. We then choose scale-free graphs as our final random graph setting to show the comparison with AF-DPOP in random graphs. Figure \ref{fig:scalefree} shows that C-CoCoA outperforms both HCMS and AF-DPOP by a significant margin.

Table \ref{tab:all} shows the comparison between C-CoCoA and HCMS on three random graph settings in terms of solution cost (C), time in sec (T) and the number of messages (M). We set the number of agents to 50 for sparse and scale-free graphs and 30 for the dense graph. Other settings are the same as the above experiments. Moreover, we stop HCMS after 500 iterations (I). C-CoCoA outperforms both HCMS and AF-DPOP in terms of solution quality, time, and number of messages in sparse and dense graphs. Note that, even after increasing the number of iterations for HCMS to 500, C-CoCoA still manages to outperform it by a significant margin (16\% for sparse graph and 8\% for the dense graph). Moreover, HCMS requires roughly 40 times more messages in the sparse graph and 200 times more messages in the dense graph than C-CoCoA. For 50 agents, AF-DPOP runs out of memory in sparse and dense settings. Thus, we omit the result of AF-DPOP for sparse and dense graph. In the scale-free setting, C-CoCoA outperforms HCMS and AF-DPOP in terms of solution quality and computation time. The closest competitor of C-CoCoA in the scale-free setting is HCMS which is outperformed by a 19\% margin in terms of solution quality and AF-DPOP is outperformed by 57\% margin roughly. Although AF-DPOP requires much less messages compared to HCMS and C-CoCoA, it requires much more time than both of these algorithms.\:It is worth noting that all the results are statistically significant for \textit{p-value \textless 0.05}.\par

\linespread{1.2}
\begin{table}[t]
\resizebox{\columnwidth}{!}{
\begin{tabular}{|c|c|c|c|c|c|}
\hline
Graph Type & Algorithm & I & C & T (sec) & M \\ \hline
\multirow{2}{*}{Sparse} & C-CoCoA & N/A & \textbf{-1266521.13} & \textbf{53.85} & \textbf{2510} \\ \cline{2-6} 
 & HCMS & 500 & -1064611.6 & 66.61 & 98464 \\ \hline
\multirow{2}{*}{Dense} & C-CoCoA & N/A & \textbf{-282012.78} & \textbf{275.66} & \textbf{2603} \\ \cline{2-6} 
 & HCMS & 500 & -260016.83 & 340.75 & 519440 \\ \hline
\multirow{3}{*}{Scale-Free} & C-CoCoA & N/A & \textbf{-713674.37} & \textbf{38.95} & 725 \\ \cline{2-6} 
 & HCMS & 500 & -575538.56 & 131.51 & 195040 \\ \cline{2-6} 
 & AF-DPOP & N/A & -306311.19 & 185.87 & \textbf{100} \\ \hline
\end{tabular}
}
\caption{Comparison between C-CoCoA and the competing algorithms in terms of Solution Cost, Time and No. of messages }
\label{tab:all}
\end{table}

\vspace{4mm}

\textbf{Random Trees:} We use the random tree configuration in our last experimental setting since the memory requirement of AF-DPOP is less on trees. The experimental configurations are similar to the random graph settings. Figure \ref{fig:tree} shows the comparison graph between C-CoCoA and the competing algorithms on random trees. The closest competitor of C-CoCoA in this setting is HCMS. On an average, C-CoCoA outperforms HCMS which in turn outperforms AF-DPOP.\par

\section{Conclusions}\label{sec:6}  
The classical DCOP model deals with discrete variables. But this assumption of the variables being discrete is not applicable to many real-world problems. Hence, the F-DCOP framework has been proposed which is a variant of DCOPs that has continuous variables. In this paper, we propose an algorithm C-CoCoA that uses Cooperative Constraint Optimization (CoCoA) technique as the underlying algorithmic framework to solve F-DCOPs. To be exact, C-CoCoA combines the discrete CoCoA algorithm with the gradient-based non-linear optimization method to solve F-DCOPs. Finally, the empirical analysis shows that C-CoCoA outperforms the state-of-the-art F-DCOP solvers, HCMS and AF-DPOP. In all the experimental settings, C-CoCoA shows better results than the other benchmarking algorithms in terms of solution quality, time, and number of message passing. In the future, we would like to further investigate the potential of C-CoCoA on various F-DCOP applications. We would also like to explore the ways to extend C-CoCoA to solve multi-objective and asymmetric F-DCOPs.

\bibliographystyle{named}
\bibliography{ijcai20}

\end{document}